\documentclass[a4paper,twoside]{article}

\usepackage{epsfig}
\usepackage{calc}
\usepackage{amssymb}
\usepackage{amstext}
\usepackage{amsmath}
\usepackage{amsthm}
\usepackage{multicol}
\usepackage{pslatex}

\usepackage{natbib}
\let\bibhang\relax

\usepackage{apalike}
\usepackage{algorithm}% http://ctan.org/pkg/algorithms
\usepackage{algpseudocode}% http://ctan.org/pkg/algorithmicx
\usepackage{subfig}
\usepackage[euler]{textgreek}
\usepackage{amssymb}
\usepackage{graphicx}
\usepackage{csquotes}
\usepackage{enumerate}
\usepackage{latexsym}
\usepackage{url}
\usepackage{subfig}
\usepackage{array}
\usepackage{spverbatim}
\usepackage{comment}
\usepackage{eurosym}
\usepackage{booktabs}
\usepackage{adjustbox}
\usepackage{subfig}
\usepackage{svg}
\usepackage{multirow}
\usepackage{adjustbox}
\usepackage{xcolor}

\usepackage{float}

\usepackage{SCITEPRESS}     % Please add other packages that you may need BEFORE the SCITEPRESS.sty package.

\begin{document}

%\title{TabooLM: A fine-tuned Language Model for Word Guessing}

\title{A Neural-Symbolic Approach Towards Identifying Grammatically Correct Sentences}

%On the effectiveness of a neural-symbolic sentence-detection tool

%A Simplified Way to Improve Language Models Performance
%\begin{comment}

\author{\authorname{Nicos Isaak\orcidAuthor{0000-0003-2353-2192}}
	\email{nicosi@acm.org}
	\affiliation{Computational Cognition Lab, Cyprus}}

%\end{comment}


\abstract{ Textual content around us is growing on a daily basis. Numerous articles are being written as we speak on online newspapers, blogs, or social media. Similarly, recent advances in the AI field, like language models or traditional classic AI approaches, are utilizing all the above to improve their learned representation to tackle NLP challenges with human-like accuracy. It is commonly accepted that it is crucial to have access to well-written text from valid sources to tackle challenges like text summarization, question-answering, machine translation, or even pronoun resolution. For instance, to summarize well, one needs to select the most important sentences in order to concatenate them to form the summary. However, what happens if we do not have access to well-formed English sentences or even non-valid sentences?
Despite the importance of having access to well-written sentences, figuring out ways to validate them is still an open area of research. To address this problem, we present a simplified way to validate English sentences through a novel neural-symbolic approach. Lately, neural-symbolic approaches have triggered an increasing interest towards tackling various NLP challenges, as they are demonstrating their effectiveness as a central component in various AI systems. Through combining Classic with Modern AI, which involves the blending of grammatical and syntactical rules with language models, we effectively tackle the Corpus of Linguistic Acceptability (COLA), a task that shows whether or not a sequence of words is an English grammatical sentence. Among others, undertaken experiments effectively show that blending symbolic and non-symbolic systems helps the former provide insights about the latter's accuracy results.}

\onecolumn \maketitle \normalsize \setcounter{footnote}{0} \vfill

\section{Introduction}
It is well-known that having access to rich textual information in order to train or build AI systems is crucial \citep{Isaak2017HowTA,kn:michael2013machines,trueswell1998prune,frazier1978sausage,stevenson2000experiments}. Numerous articles, online blogs, and other archives are being written or developed daily all over the WWW \citep{el2021automatic}. It is generally accepted that having access to grammatically correct training data allows AI researchers to build systems in order to become virtuosos of almost anything. On the contrary, grammatically incorrect data can easily lead to building systems that cannot be utilized to achieve top results, as further gains could always be achieved via a more accurate semantic analysis of the training data \citep{icaart20,ko2004improving,Gupta2010ASO,goldstein1999summarizing,stevenson2000experiments}.

To tackle challenges like text summarization, question-answering, machine translation, or even pronoun resolution, one must have access to well-written text independent of the AI method used. To illustrate, in order to summarize well, we need to select the most important sentences to concatenate them to form the summary, excluding large and short sentences \citep{Gupta2010ASO} or even grammatically wrong ones. For another, to do pronoun resolution, one needs to gather well-written sentences or even write hand-crafted syntactic or semantic rules that combine words that act as nominal subjects or direct objects \citep{Isaak2016}.
Commensurate with this, we need to consider whether a noun phrase immediately following a verb is its direct object or the subject of an embedded clause, commonly known as the garden-path effect \citep{wilson2009making}. In this regard, even the type of sentence results in an increased processing difficulty in the disambiguating region for resolving pronouns or even with abstract information regarding the grammatical properties of words in sentences, which is necessary for various NLP tasks \citep{trueswell1998prune}.

The issue that our work raises is that, regardless of the method used, having access to tools that shed light on a sentence-structure level would further enhance the building of transparent systems. On the one hand, we have opaque neural networks like the transformers' architecture or the crème de la crème of modern AI through which we can tackle almost everything \citep{vaswani2017attention,liu2019roberta,devlin2018bert}. The problem with neural networks is that they are non-transparent and, on occasions, brittle, that is, when they find something they have never met during training, they can often mess up their inferences \citep{marcus2019rebooting,mitchell2019artificial}. On the other hand, traditional Classic AI or symbolic AI systems follow a different point of view as they approach various NLP challenges through knowledge and reasoning tools that offer transparency and explainability. However, a problem these systems face is that we cannot always use simple hand-crafted rules by experts to capture the knowledge of complex environments, as they directly relate to experts' subconscious knowledge \citep{wooldridge2020road}.

In agreement with the above, here, we propose a neural-symbolic approach that combines the good of both worlds. Lately, neural-symbolic approaches have spurred interest in the research AI community as they combine the brain and mind dichotomy for added value \citep{garcez2022neural,sarker2021neuro}. In this sense, integrating both approaches in producing better tools would bring us closer to the endowment of machines with commonsense reasoning abilities like those found in humans \citep{icaart22}. Given that we cannot get to the moon by climbing successively taller trees, here, we propose a new hybrid approach by combining the best of the two worlds \citep{marcus2019rebooting}.

The aim is to utilize neural networks' densely connected collection of inferential knowledge for various NLP tasks \citep{https://doi.org/10.48550/arxiv.2010.11967}, often involving the fine-tuning on downstream tasks, and blending it with a symbolic rule-based system that utilizes expert hand-crafted rules based on semantics and pragmatics found in text. To hand-craft our rules, we use the spaCy dependency parser, through which we find existing binary relations from words in English sentences to validate and output their type, which can be simple, compound, complex, or compound-complex \citep{seely2013oxford}.

Therefore, in this study, we start by presenting what neural approaches, and especially what language models, are all about. Next, we present the symbolic approaches and shed light on how we can utilize their semantics to build transparent systems as found in \citet{Isaak2016}. In the next section, we put forward our simplified methodology of building and enhancing a rule-based system with a fine-tuned language model and continue by presenting our empirical evaluation results against the Corpus of Linguistic Acceptability (COLA), a task that shows whether or not a sequence of words is a valid English grammatical sentence. 

Undertaken experiments show that, when compared with the pure symbolic approach, the neural-symbolic blending leads to performance improvements of 10\% in accuracy. At the same time, it seems that the neural parts' accuracy results are directly related to the type of each sentence, meaning that the simplest the type of sentence, the better our results.

Our results add to the much bigger AI boom that is underway, showing that neural-symbolic approaches are able to address some of the limitations of both symbolic and non-symbolic approaches.

%(like in thinking fast and slow

\section{Language Models}
Language models are neural approaches that learn to map sequences of feature vectors to predictions so that words with similar vectors can be replaced by others, often regardless of word order and context \citep{bengio2008neural}. The language models field is indeed bustling with innovation in a way that almost every challenge we have been struggling with for many years can now be tackled with high-accuracy results. For instance, recent results have demonstrated that OpenAI's ChatGPT\footnote{\url{https://openai.com/blog/chatgpt/}}, a trained language model under reinforcement learning with human feedback (RLHF), is able to chat and have conversations with humans about numerous concepts. It can follow instructions or directions to write or debug code, give instructions about a user's current problem, or even answer questions to pass Amazon's AWS certification exam. Similarly, language models like BERT \citep{devlin2018bert}, RoBERTa \citep{liu2019roberta},  DeBERTa \citep{https://doi.org/10.48550/arxiv.2006.03654}, or OpenAI's GPT-3 \citep{brown2020language} have millions to billions of parameters that can be used for zero-shot classification, text similarity, text summarization or even pronoun disambiguation and question-answering tasks.

It is widely accepted that these models, if trained well or fine-tuned to a downstream task, can store a form of knowledge that helps them tackle challenges the good-old fashion AI (GOFAI) has struggled with for many years, in the sense that they can become \textit{a silver bullet} for almost anything. However, it is generally accepted that neural solutions are not transparent, meaning they cannot explain what led them to a specified solution which is crucial when dealing with decisions that might affect a person's life, such as loan decisions or even a court's sentence decision. Another problem is that they are limited to preprogrammed tasks, in a sense that when they face tasks for which they were never trained, or tasks involving words with similar word vectors met during training, this can often lead them to wrong conclusions. For instance, recent experiments with ChatGPT have shown that when prompted to code a well-defined problem, the model makes mistakes or even calls non-existent or non-declared variables.

%there is a long tail of completely unpredictable situations
\section{Commonsense \& Reasoning Approaches}
Since the late 50s, the AI community has been trying to tackle various problems mostly via symbolic solutions \citep{mccarthy2006proposal}. These refer to good-old fusion AI (GOFAI), which, speaking approximately, says that in order to endow machines with commonsense and reasoning abilities like those found in humans, we need to build transparent and explainable solutions.
In contrast, compared to non-symbolic or neural approaches that are often opaque and brittle, symbolic solutions, when able to reach a certain conclusion, will always be able to show their work by justifying what led them to take certain decisions \citep{marcus2019rebooting,wooldridge2020road}. 

To illustrate, MYCIN \citep{shortliffe1975model}, an expert system for diagnosing or offering bits of advice for blood diseases, was hard-coded with transparency and explainability, meaning that at any moment, it would explain its reasoning process. Additionally, rule-based systems like DENDRAL and XCON/RI were used for chemical analysis and the ordering of computer parts \citep{feigenbaum1970generality,mcdermott1980ri}. Most recently, projects like CYC,  KnowITAll, ConceptNet, Yago, Nell, Atomic, or the Websense engine have tried to solve various problems by simulating human-like reasoning abilities into machines \citep{icaart22}. 

Many of the above, like CYC or ConceptNet, try to figure out how the world works by building an extensive knowledge base of hand-coded rules from experts or crowd workers \citep{lenat1985cyc}. Others, like the KnowItAll project or the Websense engine, try to automate the building of facts, evaluated probabilistically, by utilizing resources like the WWW. However, the aforementioned commonsense and reasoning approaches seem to suffer from three drawbacks: i.) building a knowledge base from scratch is cumbersome and tedious work, ii.) being able to transfer all of the human knowledge on a piece of paper or system requires expertise, in the sense that knowing something does not mean you can also explain it in detail, iii.) human reasoning ability, while common and trivial for humans, is lacking in today’s AI systems. For example, many systems cannot induce new rules, that is, building new rules from scratch according to your knowledge base.

\section{Dependency Parsers}
Dependency parsers analyze structures of text (e.g., sentences, paragraphs) in order to discover the syntactic dependencies on a word level, meaning binary asymmetric relations the words are linked \citep{kubler2009dependency,rasooli2015yara}. As stated by  \citet{kubler2009dependency}, each word found in a sentence is de facto connected to other words in a way that not only is not isolated, but it plays a significant role in the sentence structure. On another, each connection of words refers to binary relations between superior and inferior words, that is, words that have grammatical functions based on other words in a sentence \citep{nivre2005dependency}. 

To illustrate, we can have relations such us nsubj (cat, plays) and dobj (plays, ball), where the former refers to the nominal subject and the latter to the direct object of a sentence. Binary relations can be further combined to develop enhanced relationships, called scenes, such as cat (1), ball (2), plays (1, 2), indirectly telling us that a \textit{cat is playing with a ball} \citep{Isaak2016,michael2013machines}.   

On a second front, via scenes like the above, we can tap into the richness of the knowledge and reasoning area to tackle challenges like the Winograd Schema Challenge \citep{levesque2014our} or the Taboo Challenge Competition \citep{isaak2023zero,inproceedingsTabooIsaak}. In this sense, utilizing dependency parsers could lead to the development of systems able at finding task-specific correlations and building convoluted relationships between words in sentences to tackle various NLP tasks.

Given the difficulties of having pure opaque non-symbolic systems or transparent ones that cannot easily induce new rules to build their knowledge, below we present a new neural-symbolic approach to identify grammatically correct English sentences. As this is still an open area of research, we start with our methodology that blends the good of both worlds, continue with our experimental section, where we present our results, and finally summarize with our conclusion section.

%According to \citet{Radford2019LanguageMA}, when trained on large or a variety of datasets, language models start learning various relations between sequences of text without requiring explicit supervision. In this section, we are examining if additional fine-tuning via dependency parsers, lead language models to achieve better results on various NLP tasks.

\section{Methodology}
We start by presenting our approach based on its various components (see Figure \ref{methodologyfig}). At its hard, there are two main parts consisting of various components, namely the symbolic and the neural one (displayed as dashed rectangles in Figure \ref{methodologyfig}). For concreteness, when referring to our neural-symbolic approach, we refer to all the components above that are being utilized to validate English sentences. Additionally, when referring to parsers, we refer to dependency parsers that analyze text sequences to perform word-level analysis based on binary relations found between the words \citep{kubler2009dependency}.

\subsection{The Symbolic Part}
\begin{figure*}[h!]
	\centerline{\includegraphics[width=\textwidth]{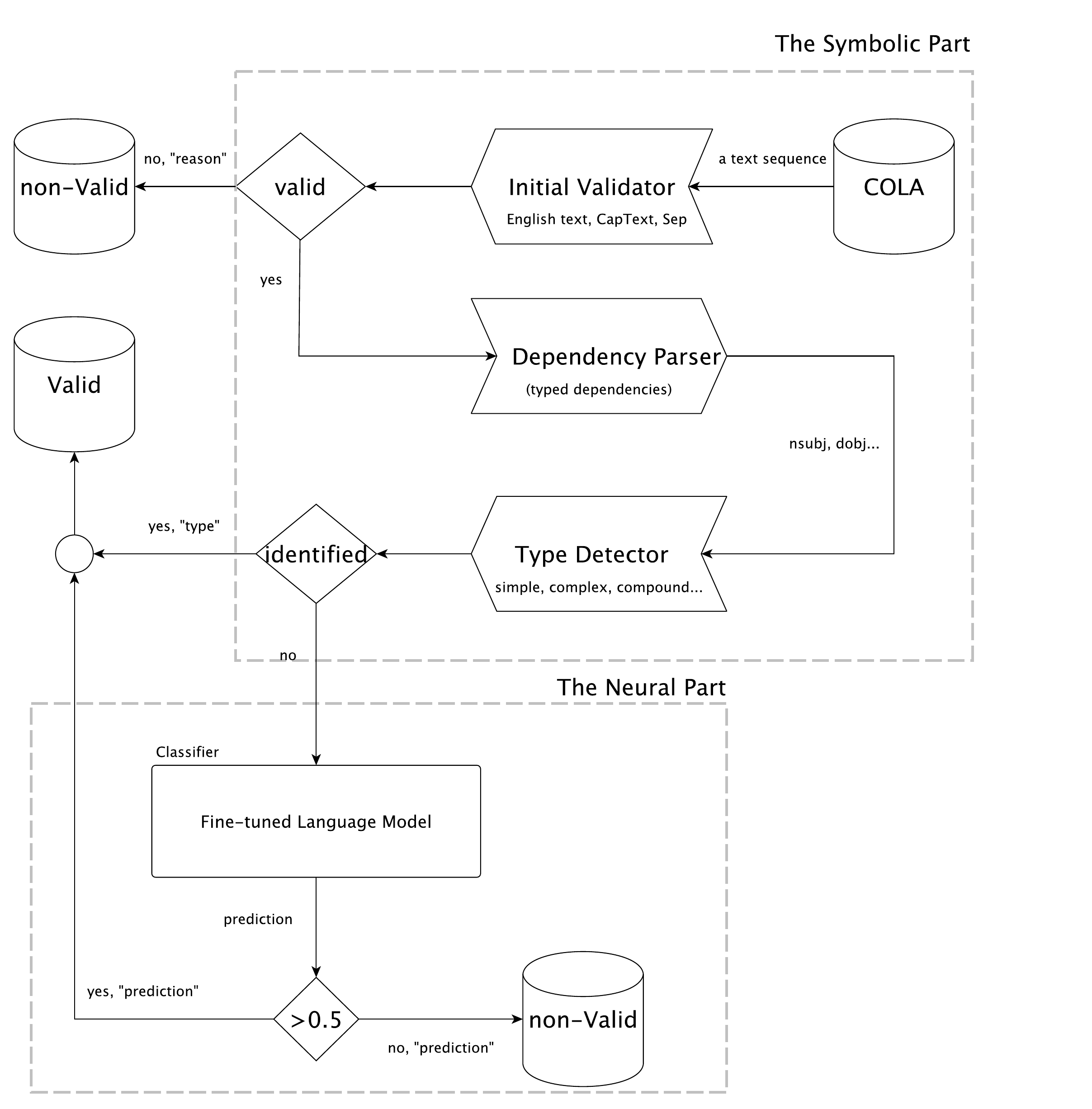}}
	\caption{Our Neural-symbolic Approach to Identify Grammatically-Correct English Sentences: The symbolic components are displayed in the upper-right corner, while the neural ones are displayed in the down-left corner.}
	\label{methodologyfig}
\end{figure*}

The symbolic part consists of transparent components that show their work at every step of the validation process. As shown in Figure \ref{methodologyfig}, the whole process starts with the \textit{Initial Validator}, responsible for filtering out sentences that do not follow the basic sentence pattern, that is, starting with a capital letter and ending with a declarative separator. From this point on, the sentence validation process is, to all intents and purposes, a rule-based approach, which is defined in the following paragraphs. 

\begin{figure*}[h!]
	\centerline{\includegraphics[width=\columnwidth]{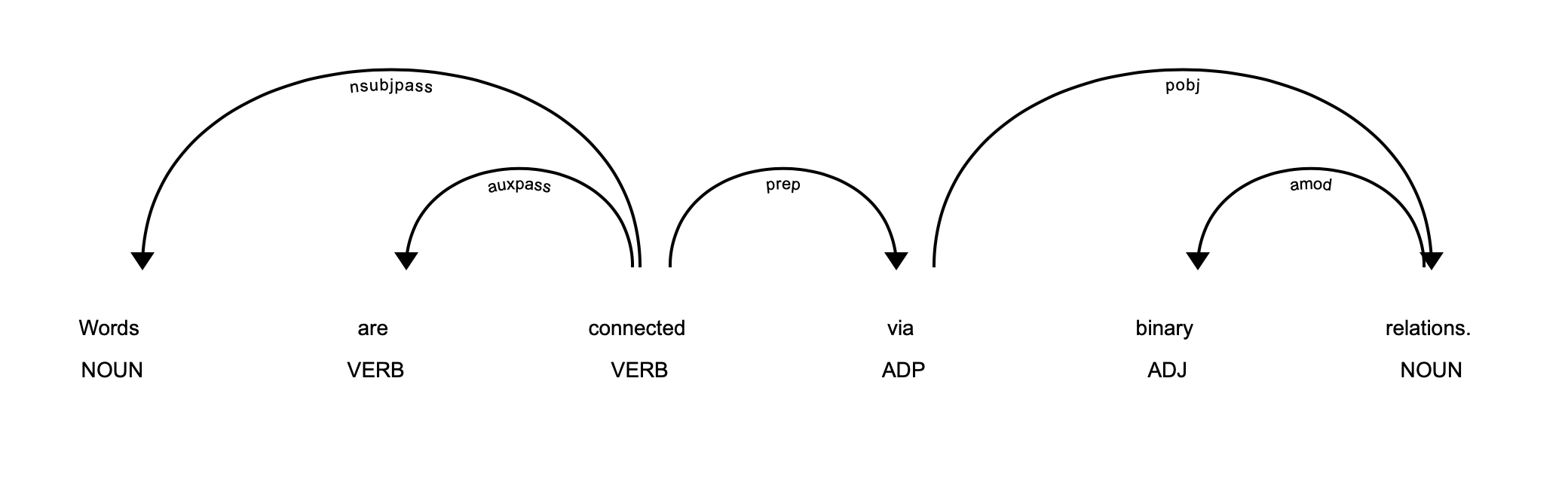}}
	\caption{An example of a Dependency Parser Output: Parsers can analyze sentences and display binary relations between words.}
	\label{depgraphexample}
\end{figure*}

Every first-stepped validated sentence is then parsed by the spaCy dependency parser\footnote{https://spacy.io} (see dependency parser in Figure \ref{methodologyfig}). The whole idea behind parsing is that each sentence consists, or better, it is built upon binary asymmetric relation between its words (see Figure \ref{depgraphexample}). On another, it is well-known that words or even groups of them play various grammatical roles, depending on their position in a sentence or even their part of speech \citep{nivre2005dependency}. According to \citet{Isaak2016}, in a further step, we can analyze these relations to extend them in another abstract layer of relations, which are scenes between superior and inferior words found in sentences ---for instance, through scenes, we can relate subjects and objects through verbs they are connected with.

Following, each and every typed dependency is given to the \textit{Type Detector} component, which is responsible for identifying whether a sentence is of simple, compound, complex, or compound-complex type. The general idea behind the \textit{Type Detector} consists of the following steps: 
\begin{itemize}
	\item For any given sentence, start the process by identifying its number of subjects and objects. 
	\item If no subjects or objects exist, consider this a non-valid English sentence. 
	\item Check some given \textit{flags} to identify whether the sentence can be a compound, complex, or compound-complex one. If common flags were found, then this is a compound-complex sentence. However, if the returned flags could only be found in compound sentences, this is a compound sentence, otherwise, this is a complex one.
	\item If nothing could be identified, but at least one subject exists, consider this a simple sentence.
\end{itemize}

Figure \ref{meth-typeDetectorfig} shows the whole procedure behind the \textit{Type Detector} component in a lucid and approachable way. Given that English is a Subject-Verb-Object language, for any given sequence of text, we start by utilizing the spaCy dependency parser to output a required number of typed dependencies, such as nominal subjects, direct objects, compounds, objects of prepositions, or indirect objects. If at least one object and subject exist, we proceed to the next step, whereas otherwise, the \textit{Type Detector} returns a \textit{non-valid} message, which, speaking approximately, it notifies us of a non-valid sentence ---this is the exact point where the \textit{neural} mechanism begins its part.

From that point on, if the total number of subjects and objects equals one, these point towards a \textit{simple} sentence. Otherwise, depending on their number, this might point to a sentence of any other type. Given that there are situations that naturally call for reasoning, as it is hard even for humans to identify between complex compounds and complex-compound sentences, here we are employing a simple technique that focuses on the type of several connectors that different parts of a sentence connect with. 

To represent the connectors at hand, we group them into two sets of identifiers: compound and complex (see Compound and  Complex identifiers in Figure \ref{meth-typeDetectorfig}) \citep{seely2013oxford}. The former ones refer to connectors that usually connect parts of compound sentences, meaning sentences grouped by independent clauses connected by the well-known FANBOYS [``, for ", ``, and ", ``, nor ", ``, but ", ``, or ", ``, yet ", ``, so "] or other connectors like [``; however,", ``; moreover,", ``; nevertheless,", ``; nonetheless,", ``; therefore,", ``; but"]. The latter group refers to other miscellaneous connectors such as [``because ",``since ",``so that ", ``although ", ``even though ", ``though ", ``whereas ", ``while ", ``where ", ``wherever ", ``how ", ``however ", ``if ", ``whether ", ``unless ", ``that ", ``which ", ``who ", ``whom ", ``after ", ``as ", ``before ", ``since ", ``when ", ``whenever ", ``until "], mostly used in complex sentences in order to connect dependent with independent clauses. The variability of the connectors used generally stems from the type of sentence at hand, as it is unclear where, how, under what circumstances, or by whom it was designed. From that point on, if both types of connectors are used, this points to a compound-complex sentence. 

However, given that a specified number of words could lead to an unlimited number of sentences \citep{adger2019language} and that every language is an evolving mechanism humans use to communicate, all the above is not a panacea. On another, humans might use innovative ways to connect clauses in sentences, leading us to our proposed engine's neural part.

\subsection{The Neural Part}
Given that our symbolic part cannot fully replicate the cognitive mechanisms used by humans when constructing sentences, this points forward to our neural part. This is nonetheless a non-transparent fine-tuned language model on a downstream task that validates sentences. 

\begin{figure*}[h!]
	\centerline{\includegraphics[width=\textwidth]{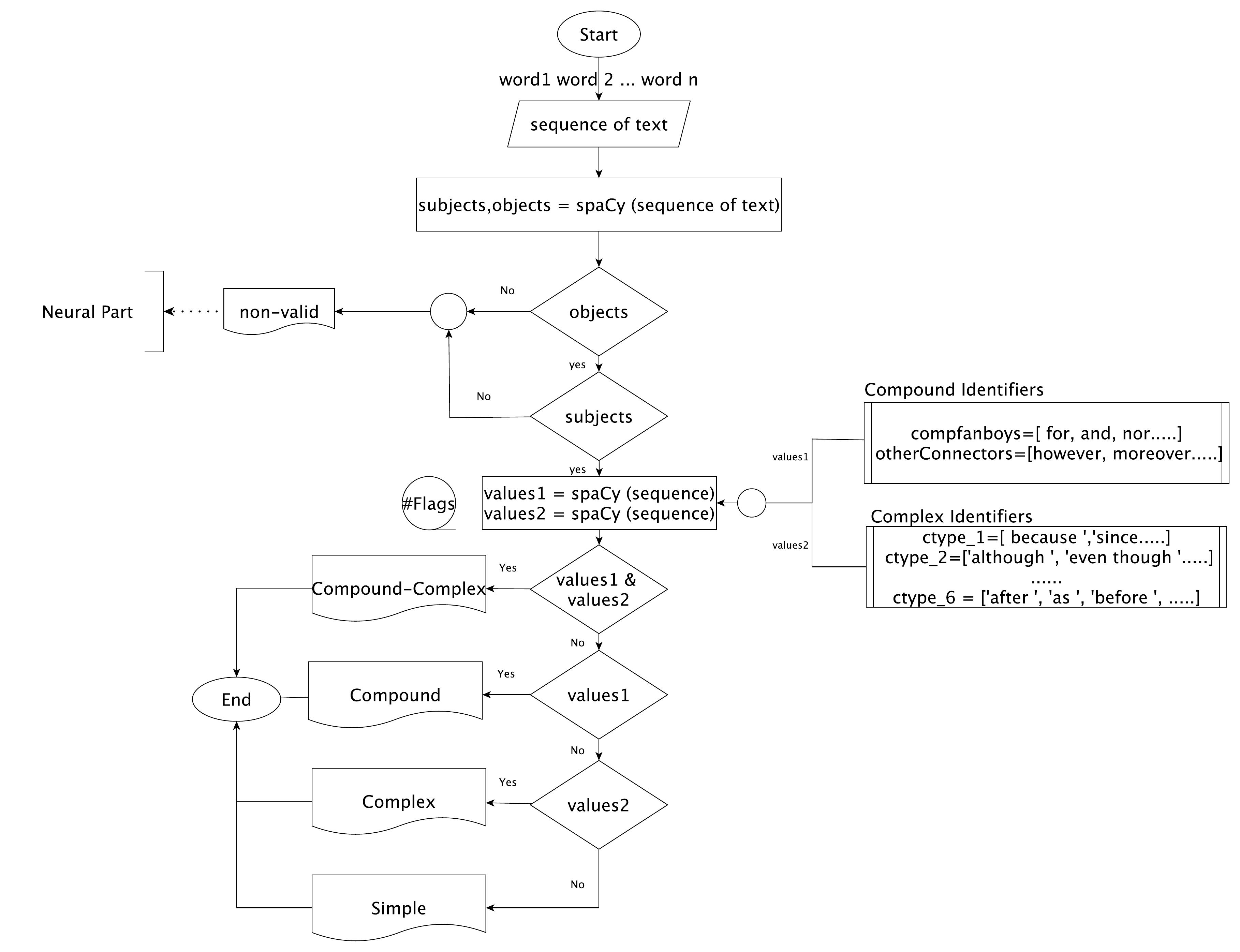}}
	\caption{An Extensive Analysis of the \textit{Typed-Detector} Component: Given a sequence of text, it outputs whether this is a simple, compound, complex, or compound-complex sentence. If it cannot determine its type, it notifies us with the unknown flag.}
	\label{meth-typeDetectorfig}
\end{figure*}

Even though language models often cannot explain their work, their densely connected networks can capture long-term relations between sequences of words to tell us whether or not we have valid sentences. In this regard, the neural part, responsible for identifying grammatically written sentences, is called out either when our symbolic part cannot categorize a sentence or the dependency parser cannot detect any subjects or objects (see Figure \ref{meth-typeDetectorfig}). 

The Neural Part is built upon the well-known BERT model \citet{devlin2018bert}, one of the first transformer models built in 2018. The Bert model is so important that the term \textit{Bertology} was coined to refer to models that employ similar mechanisms. Specifically, Bert and its kin are everywhere, from sentiment analysis to pronoun resolution tasks helping at the same time research groups to tackle various challenges \citep{10.1145/3442188.3445922}.

Given that language models like BERT are acting like stochastic parrots that stitch together various parts of our language \citep{10.1145/3442188.3445922} without any communicative intent, further fine-tuning is needed as an optimal way to train our model for our downstream task. To that end, the COLA dataset is utilized, based on which our enhanced neural model can detect whether or not a sequence or text is a valid English sentence\footnote{\url{https://huggingface.co/datasets}}.

\begin{table*}[!h]
	\caption{A snapshot of two training examples from COLA dataset. Given the text, we calculate whether or not this is a valid sentence. 
			\label{yelpreviewexample}} \centering
		\begin{adjustbox}{max width=\textwidth}%{max width=\textwidth}
			\begin{tabular}{|c|c|}
				\hline 
				Sentence & Label \tabularnewline
				\hline 
				\hline 
				The professor talked us into a stupor. & 1\tabularnewline
				\hline 
				The professor talked us. & 0\tabularnewline
				\hline 
			\end{tabular}

		\end{adjustbox}
	\end{table*}

\begin{table*}[htb!]
	\caption{Experimental Evaluation Results: Our NeSy (Neural-Symbolic) approach achieves 72\% accuracy in identifying grammatically correct sentences from the COLA dataset. 
		\label{resultsCola}} \centering
	\begin{adjustbox}{max width=\textwidth}%{max width=\textwidth}
		% Preview source code for paragraph 0
		
		\begin{tabular}{|c|c|c|c|c|c|c|}
			\hline 
			Platform & Unknown & Simple & Compound & Complex & Compound-Complex & \% Correct \tabularnewline
			\hline 
			\hline 
			Neural-Symbolic  & 76 & 270 & 41 & 116 & 24 & 72\%\tabularnewline
			\hline 
			Neural & 76 & - & - & - & - & 83\%\tabularnewline
			\hline 
			Symbolic & - & 270 & 41 & 116 & 24 & 66\%\tabularnewline
			\hline 
		\end{tabular}

	\end{adjustbox}
\end{table*}

\begin{table*}[h!]
	\caption{Experimental Evaluation Results: Based on the various types of sentences identified by our symbolic part, the neural part correctly marked as valid or not 87\% of the simple, 83\% of the compound, 73\% of the complex, and 58\% of the compound-complex sentences.  
		\label{resultsColaNeuralPart}} \centering
	\begin{adjustbox}{max width=\textwidth}%{max width=\textwidth}
		\begin{tabular}{|c|c|c|}
			\hline 
			\multirow{2}{*}{} & \multirow{2}{*}{Number of Identified Sentences} & \multirow{2}{*}{Correctly Identified (Neural Part)}\tabularnewline
			&  & \tabularnewline
			\hline 
			\hline 
			Simple  & 270 & 87\%\tabularnewline
			\hline 
			Compound & 41 & 83\%\tabularnewline
			\hline 
			Complex & 116 & 73\%\tabularnewline
			\hline 
			Compound-Complex & 24 & 58\%\tabularnewline
			\hline 
		\end{tabular}
	\end{adjustbox}
\end{table*}

\section{Experimental Evaluation}%Experimental Design and Results
Here, we illustrate how well our neural-symbolic (NeSy) engine identifies grammatically correct English sentences. For our experiments, we used a subset of the COLA dataset consisting of two sets, a testing one of 527 and a training one of 8551 examples responsible for fine-tuning the neural part\footnote{https://huggingface.co/datasets/linxinyuan/cola}. For training purposes of our neural part, we utilized a dedicated GPU from the Gradient Paperspace platform\footnote{https://www.paperspace.com}. After the training procedure, and based on our methodology, we stepped into the testing phase, summarized in two steps: i) For each examined sequence of text, the symbolic part, which is responsible for identifying its subjects and objects, utilizes some connectors to categorize it as a simple, compound, complex, or a compound-complex sentence. ii) if no objects, subjects, or even connectors are found, the neural part is called out to identify, based on its densely connected network of relations, if the sequence of text at hand is a valid English sentence.

\subsection{Results}
According to our results, summarized in Table \ref{resultsCola}, the NeSy approach achieved an accuracy of 72\% in identifying whether a text sequence is a grammatically correct English sentence. At the same time, the symbolic part categorized 86\% of the examined sentences, leaving the rest 14\% to be exclusively resolved by the neural part. As stated in Table \ref{resultsCola}, among 451 sentences, the symbolic part categorized 270 as simple, 41 as compound, 116 as complex, and 24 as compound-complex sentences.

An undertaken abbreviation study showed that the symbolic part could correctly categorize 66\% of the examined sentences either as valid or invalid ones. A text sequence was considered an invalid sentence when the symbolic part could not determine its type. At the same time, the neural part can correctly validate 83\% of the 76 sentences the symbolic part could not determine their type. It is worth noticing that the neural part, referring to the fine-tuned BERT model, when running solely on the entire testing set, achieves an accuracy of 82\%, which might indirectly show that sentences the symbolic approach could not resolve were relatively easier for the neural part.

One might argue that we can achieve high-accuracy results with the neural part alone, which outperforms the symbolic one by a large margin of 16\% in accuracy. However, regardless of the results, we must keep in mind that the neural part is opaque and brittle, acting like a stochastic parrot that cannot guarantee the required consistency in achieving the same results. Alternatively, the symbolic part's inner mechanisms are always the same, meaning that it will always be able to achieve almost the same results, which, as far as we can determine, will be around the same percentage of accuracy. Of course, semantic and pragmatic issues following every person's cognitive behavior upon designing sentences or phrases cannot determine that we will always be able to capture its inner connections to output or validate their type, which brings us directly to our implemented neural-symbolic approach. Nonetheless, figuring out ways to validate sentences is still an open area of research that needs further attention from the research community.

Recent experiments revealed a direct relation between the type of sentence and the effort it took for a system to output its hardness, meaning its ability to resolve pronouns in sentences of various types \citep{GCAI-2018:Data_Driven_Metric_of_Hardness,10.1093/logcom/exab005}. The implicit assumption here was that the more challenging the sentence is, the more difficult it is for our neural part to correctly categorize it as valid or non-valid. In this regard, a sentence-type analysis we undertook revealed a direct relation between each sentence type and our neural part's ability to validate it. Specifically, the neural part correctly validates 87\% simple, 83\% compound, 73\% complex, and 58\% compound-complex sentences. The results ultimately show that the neural part's accuracy results are directly related to each sentence type, showing the importance of our symbolic part and its role in our neural-symbolic engine. It thus seems that having insights regarding neural net performance is crucial when dealing with semantic or even pragmatic issues. For the task at hand, neural nets' ability generally stems from factors like each sentence's subjects, objects, and clauses and how all these connect. Overall, the results show the necessity of both our symbolic and non-symbolic parts, making our hybrid approach the optimal way to deliver the most accurate results.

\section{Conclusion}
We have demonstrated the plausibility of integrating a neural-symbolic approach to identify grammatically correct sentences. Despite the importance of utilizing well-written sentences in a plethora of natural language tasks, there has been little research into building systems that perform even the simplest task of deciding on a sentence's validity, whether a sentence is grammatically correct or not.

Given a number of words, our system determines whether or not this is a grammatical English sentence. At the same time, through a symbolic component, it generates its type, whether this is a simple, complex, compound, or complex-compound sentence. A fine-tuned language model can also be called to output a confidence score in the range of 0 to 1, where the higher the value, the higher the confidence for an English sentence. We consider this neural-symbolic property advantageous for four reasons. First, the system starts the sentence validation process through the symbolic component, a transparent non-opaque way that tells users why a certain sentence is valid or not. Second, given that neural components are opaque and sometimes brittle, our neural component is called a limited number of times, and that is when the symbolic component cannot determine a sentence's type. Third, the neural component is upgradeable, meaning we can easily replace the BERT model with any other model we want to fine-tune on this task. Finally, it can offer insights about each sentence type, which is directly related to our neural part's accuracy results.

As building systems that discern a sentence's grammatical correctness is still challenging, we hope this work will act as a simple proxy to spur further research on this problem to explore and offer more fine-grained strategies. We hope this work might be the starting point for a new slew of applications combining neural nets with symbolic ones to identify grammatically correct sentences.

For the foreseeable future, researchers can build on this work by enhancing its capabilities in various ways, such as building a better symbolic mechanism or even blending the two components further. For example, one could incorporate a neural network to identify each sentence's type to add to the symbolic component's decisions.

%\clearpage
\bibliographystyle{apalike}
{\small
	\bibliography{isaak}}

\end{document}